\documentclass[AMA,Times1COL]{WileyNJDv5} 

\articletype{Article Type}%

\received{Date Month Year}
\revised{Date Month Year}
\accepted{Date Month Year}
\journal{Journal}
\volume{00}
\copyyear{2023}
\startpage{1}

\usepackage{xcolor}
\usepackage{makecell}
\usepackage{colortbl}
\definecolor{yellow}{rgb}{1, 1, 0.7}
\definecolor{orange}{rgb}{1, 0.85, 0.7}
\definecolor{tablered}{rgb}{1, 0.7, 0.7}

\newcommand{\bc}{\mathbf{c}}

\newcommand{\bJ}{\mathbf{J}}

\newcommand{\bp}{\mathbf{p}}

\newcommand{\bR}{\mathbf{R}}
\newcommand{\bS}{\mathbf{S}}

\newcommand{\bW}{\mathbf{W}}
\newcommand{\bx}{\mathbf{x}}

\newcommand{\cG}{\mathcal{G}}

\newcommand{\bmu}{\boldsymbol{\mu}}

\newcommand{\bSigma}{\boldsymbol{\Sigma}}

\begin{document}

\title{Efficient 3D Scene Reconstruction and Simulation from Sparse Endoscopic Views}

\author[1]{Zhenya Yang}



\authormark{Zhenya Yang}
\titlemark{Efficient 3D Scene Reconstruction and Simulation from Sparse Endoscopic Views}

\address[1]{The Chinese University of Hong Kong}



\corres{\email{zyyang673@gmail.com}}



\abstract[Abstract]{
Surgical simulation is essential for medical training, enabling practitioners to develop crucial skills in a risk-free environment while improving patient safety and surgical outcomes. 
However, conventional methods for building simulation environments are cumbersome, time-consuming, and difficult to scale, often resulting in poor details and unrealistic simulations.
In this paper, we propose a Gaussian Splatting-based framework to directly reconstruct interactive surgical scenes from endoscopic data while ensuring efficiency, rendering quality, and realism. 
A key challenge in this data-driven simulation paradigm is the restricted movement of endoscopic cameras, which limits viewpoint diversity.
As a result, the Gaussian Splatting representation overfits specific perspectives, leading to reduced geometric accuracy.
To address this issue, we introduce a novel virtual camera-based regularization method that adaptively samples virtual viewpoints around the scene and incorporates them into the optimization process to mitigate overfitting.
An effective depth-based regularization is applied to both real and virtual views to further refine the scene geometry.
To enable fast deformation simulation, we propose a sparse control node-based Material Point Method, which integrates physical properties into the reconstructed scene while significantly reducing computational costs.
Experimental results on representative surgical data demonstrate that our method can efficiently reconstruct and simulate surgical scenes from sparse endoscopic views.
Notably, our method takes only a few minutes to reconstruct the surgical scene and is able to produce physically plausible deformations in real-time with user-defined interactions. 
}


\jnlcitation{\cname{%
\author{Taylor M.},
\author{Lauritzen P},
\author{Erath C}, and
\author{Mittal R}}.
\ctitle{On simplifying ‘incremental remap’-based transport schemes.} \cjournal{\it J Comput Phys.} \cvol{2021;00(00):1--18}.}

\maketitle

\renewcommand\thefootnote{}

\renewcommand\thefootnote{\fnsymbol{footnote}}
\setcounter{footnote}{1}

\section{Introduction}\label{sec:intro}

Realistic and efficient endoscopic scene simulation is crucial for surgical training, education, and surgical embodied intelligence.
Existing approaches~\cite{hirota2003improved,bar2006simbionix,sofa} for simulating deformable objects in anatomical scenes depend on manually designed 3D models and textures, which often fail to capture the realistic appearance of various tissues and endoscopic illumination. 
Moreover, the results from these simulators show a significant gap from real endoscopy scenes and lack diversity, limiting their utility for surgical robot learning.
Recent advancements in 3D reconstruction techniques~\cite{kerbl20233d, wang2024dust3r, wang2025vggt} have sparked interest in developing a realistic, efficient, and data-driven surgical scene simulation pipeline: \emph{can we perform realistic and efficient simulation on 3D scenes reconstructed from real-world endoscopic images?} To achieve this real-to-sim transformation in the surgical field, a feasible approach is to reconstruct a 3D representation of a surgical scene from real-world data and then efficiently simulate it. However, this remains a significant challenge.

As the first part of this pipeline, reconstructing 3D endoscopic scenes from sparse views remains challenging.
Previous works introduce NeRF \cite{wang2022neural,zha2023endosurf,yang2023neural} and 3D Gaussian Splatting (3DGS) \cite{zhu2024endogs,liu2024endogaussian,huang2024endo4dgs} to reconstruct dynamic surgical scenes from endoscopy videos. 
These methods utilize neural networks to model tissue action and deformation over time, while our pipeline aims to first reconstruct a static 3D endoscopic scene from sparse views and then embed physics into the reconstructed scene for controllable deformation simulation.
The key difference is that the aforementioned methods can only fit deformation in surgical videos, whereas our method can generate realistic deformation based on user-defined interaction, making it more generalizable for downstream applications like surgeon training and simulation environments for surgical robot learning.
Due to its fast training speed and real-time rendering capability, we utilize 3DGS\cite{kerbl20233d} as the foundation of our method. However, vanilla 3DGS tends to overfit in sparse view reconstruction. 
To solve this problem, FSGS \cite{fsgs} proposes Gaussian Unpooling to guide Gaussian densification with sparse views and synthesizes pseudo views for regularization. 
MVSGaussian \cite{mvsgs} designs a generalizable model to initially reconstruct a reasonable 3D Gaussian representation, followed by per-scene optimization. 
In this paper, we propose an effective geometry regularization strategy utilizing virtual cameras to address endoscopic sparse view reconstruction.

Meanwhile, the efficient simulation of the reconstructed 3D endoscopic scene remains unsolved. PhysGaussian\cite{xie2023physgaussian} introduces physical simulation into 3DGS by deforming 3D Gaussians using deformation gradients of the Material Point Method (MPM)\cite{mls-mpm}. 
However, its simulation speed becomes extremely slow with a large number of Gaussians.
SimEndoGS~\cite{simendogs} is the first to adopt PhysGaussian in the surgical domain, but it also inherits its limitations.
To accelerate simulation, VR-GS \cite{vrgs} deforms 3D Gaussians using the mesh deformation gradient, but it requires a closed mesh.
However, endoscopic scene reconstruction only captures the surface, necessitating additional processing before simulation.
Fast-EndoNeRF \cite{fastendonerf} proposes extracting closed meshes from the radiance field and then sending them to NVIDIA Isaac Sim\cite{mittal2023orbit}. 
Nevertheless, its simulation quality is hindered by the loss of illumination and material properties during mesh extraction.
Inspired by the \textit{Sparse-Controlled Gaussian Splatting} (SCGS)\cite{SCGS}, we enhance PhysGaussian by employing sparse control nodes, greatly reducing the number of material points that need to be simulated, thereby increasing simulation speed. 
Experimental results demonstrate that our method achieves realistic physical simulation in real time without the need for complex geometry processing.

In this paper, we endeavor to simulate reconstructed surgical scenes captured from sparse endoscopic images in a completely data-driven manner. We summarize our contributions as follows: 
\textbf{1)} A robust reconstruction method is proposed to reconstruct reasonable 3D endoscopic scenes from very sparse endoscopic views. 
\textbf{2)} An efficient scene simulation method based on sparse control nodes is proposed to greatly speed up soft tissue simulation on the reconstructed endoscopic scenes. 
\textbf{3)} An efficient and automated pipeline that combines the two parts mentioned above. 
Experimental results on robotic surgery datasets demonstrate that our method can support soft-tissue interactions with realistic deformation in real-time.

\section{Method}

\begin{figure}[t]
    \includegraphics[width=\textwidth]{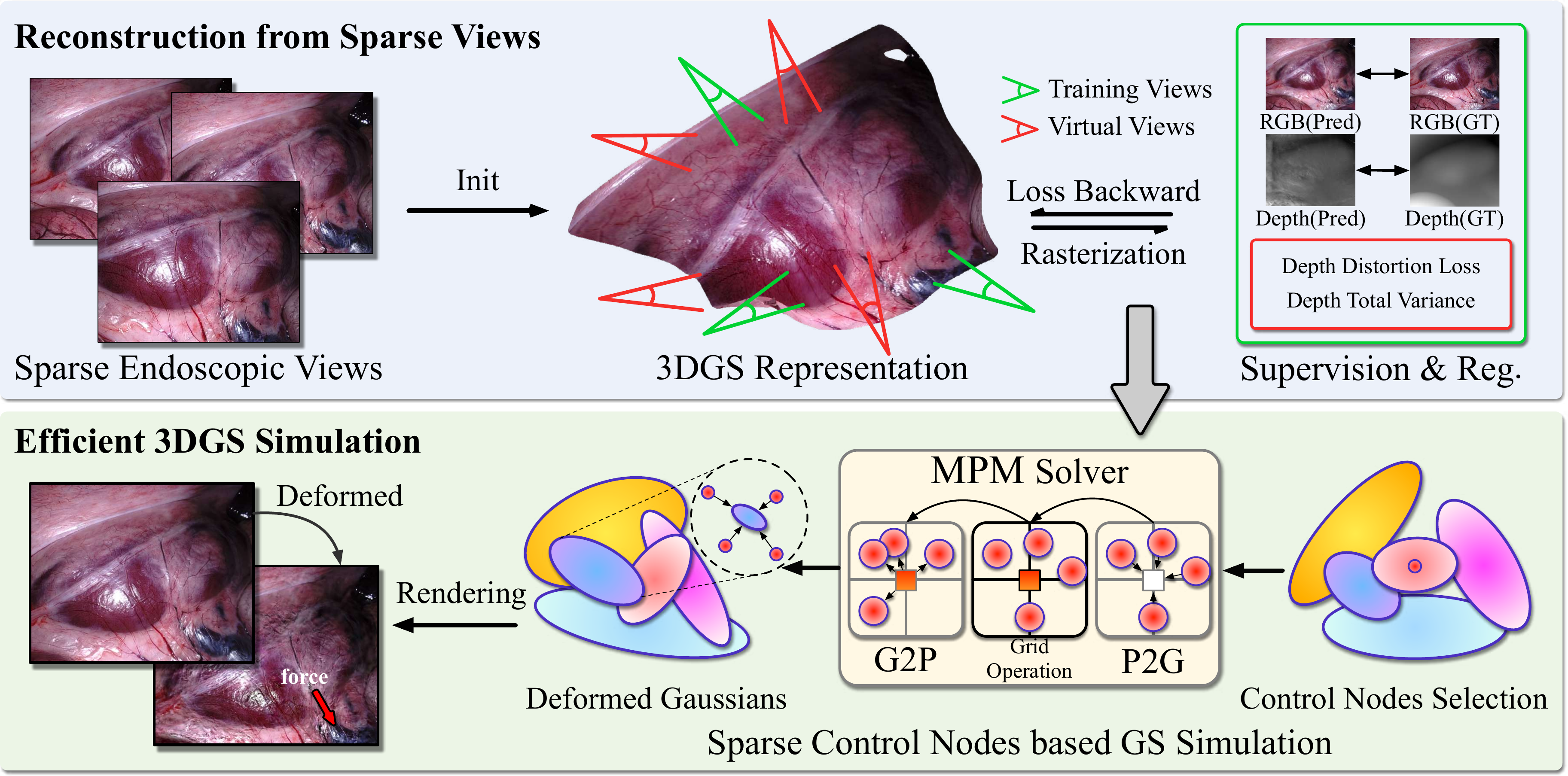}
    \caption{
        {\bf An overview of the proposed data-driven surgical simulation framework.}
        It consists of automatic scene reconstruction and physically-based scene simulation using 3D Gaussians.
        The reconstruction uses virtual camera-based depth regularization for high-quality results from sparse endoscopic views. The simulation selects sparse control nodes to represent the entire 3D endoscopy scene, significantly reducing the computation workload of simulation.
        }
    \label{overview}
\end{figure}

Provided with several sparse endoscopic views, we aim to develop a fully automatic framework to reconstruct 3D endoscopic scenes using these views and perform efficient and physically-based soft tissue simulation on the reconstructed scenes. 
We firstly give a preliminary of 3DGS in Section \ref{subsec:perliminary}. Then we improve 3D Gaussian Splatting with depth from a diffusion prior and geometry regularization in Section \ref{sec regular} to overcome the overfitting caused by sparse views. To further address this issue, we propose virtual camera-based regularization in Section \ref{sec virtual} to enforce reasonable geometric structures beyond the training views.
Based on surgical scenes reconstructed from endoscopic images, we develop an efficient 3DGS simulation method in Section \ref{sec simulation} for real-time, physically-based soft tissue simulation. 
Fig.~\ref{overview} shows an overview of our proposed reconstruction and simulation framework.
 
\subsection{Preliminary of 3D Gaussian Splatting}
\label{subsec:perliminary}
Kerbl~\etal~\cite{kerbl20233d} utilize learnable 3D Gaussian primitives to represent 3D scenes and render different views using a differentiable volume splatting rasterizer. In 3DGS, each 3D Gaussian primitive is parameterized using a 3D covariance matrix $\bSigma$ and a distribution center $\bmu$:
\begin{equation}
\cG(\bx) = \exp\left(-\frac{1}{2} (\bx - \bmu)^\top \bSigma^{-1}(\bx - \bmu)\right)
\label{eq:gaussian}
\end{equation}
During optimization, the covariance matrix $\bSigma$ is factorized into a scaling matrix $\bS$ and a rotation matrix $\bR$ as $\bSigma = \bR \bS \bS^\top \bR^\top$ to ensure its positive semidefiniteness.
To obtain the rendering results of 3D Gaussians from a specific view, the 3D Gaussian is first projected to a 2D splat in screen space using the view matrix $\bW$ and an affine approximated projection matrix $\bJ$ as illustrated in \cite{zwicker2001ewa}:
\begin{equation}
\bSigma' = \bJ \bW \bSigma \bW^\top \bJ^\top
\end{equation}
By removing the third row and column of $\bSigma'$, we obtain a $2 \times 2$ matrix, which represents the covariance matrix $\bSigma^{2D}$ of the 2D splat $\cG^{2D}$. Finally, the color of each pixel $\bp$ can be computed using volumetric alpha blending as follows:
\begin{equation}
\bc(\bp) = \sum_{i=1}^K \bc_i \alpha_i \prod_{j=1}^{i-1} (1 - \alpha_j),\quad
\alpha_i = \sigma_i \cG^{2D}_i(\bp)
\end{equation}
where $i$ is the index of the Gaussian primitives covering the current pixel, $\sigma_i$ denotes the opacity values, and $\mathbf{c}_i$ represents the view-dependent appearance modeled using Spherical Harmonics. All attributes of the 3D Gaussian primitives $(\boldsymbol{\mu}, \boldsymbol{\Sigma}, \boldsymbol{c}, \sigma)$ are optimized using the photometric loss between the rendered images and the ground-truth images.

\subsection{Regularized 3DGS for Sparse Views Reconstruction}
\label{sec regular}
We represent endoscopic scenes as a group of 3D Gaussians, each parameterized by $(\boldsymbol{\mu}, \boldsymbol{\Sigma}, \boldsymbol{c}, \sigma)$, corresponding to position, covariance matrix, color, and opacity, respectively. Optimizing all parameters of 3DGS using sparse endoscopic views can lead to overfitting. To address this, we leverage a foundation monocular depth estimation model, Depth Anything-V2\cite{depth_anything_v2}, to predict depth maps for the sparse endoscopic images. 
We incorporate the L1 loss between the estimated and rendered depth maps into our objective function to supervise scene geometry:
\begin{equation}
    \mathcal{L}_{depth}(\mathbf{p})=\mathcal{L}_{1}(\boldsymbol{D}(\mathbf{p}), \sum_{i=1}^K d_i \alpha_i \prod_{j=1}^{i-1} (1-\alpha_j))
\end{equation}
where $d_i$ is the z-value of the i-th Gaussian and $\boldsymbol{D}$ is the estimated depth map. 
However, rendered depth supervision is insufficient as GS rasterization ignores the distance between splatted Gaussian primitives.
This may result in excessive distances between Gaussians, leading to poor geometric quality.
To mitigate this problem, we introduce a depth distortion loss\cite{2dgs} to minimize the intervals between Gaussians in the splatting direction, computed as follows:
\begin{equation}
    \mathcal{L}_{dist} = \sum_{i,j}w_iw_j|z_i - z_j|,
\end{equation}
where $w_i$ and $z_i$ are the blending weight and depth of the i-th Gaussian. 
We notice that 3DGS may produce unreasonable Gaussians to fit the targeted depth map, leading to unnatural sharp fringes. 
To mitigate this, we compute the Total Variance (TV) of the rendered depth map and add this term to the objective function. 
Thus, our whole objective function can be expressed as:
\begin{equation}
    \mathcal{L} = \mathcal{L}_{GS} + \lambda_{depth}\mathcal{L}_{depth} + \lambda_{dist}\mathcal{L}_{dist} + \lambda_{TV}\mathcal{L}_{TV},
    \label{eq:train_loss}
\end{equation}
where $\lambda_{depth}$, $\lambda_{dist}$, and $\lambda_{TV}$ are hyperparameters controlling the degree of regularization, and $\mathcal{L}_{GS}$ is the loss function of the vanilla 3DGS. Our experimental results demonstrate that this regularization strategy effectively reconstructs reasonable geometry of endoscopic scenes from very sparse inputs. 

\subsection{Virtual Camera Based Depth Regularization}
\label{sec virtual}
Relying solely on the aforementioned depth supervision and regularization cannot ensure a reasonable geometric structure, as the rendered depth may also overfit the training views.
To tackle this challenge, 
we propose a virtual camera-based depth regularization strategy to refine the geometry of reconstructed scenes from multiple virtual views. 
During training, we create virtual cameras from the two closest training views using:
\begin{equation}
(q_v, t_v)_\alpha = \left(\text{Slerp}\left(q_1, q_2, \alpha\right), \text{lerp}\left(t_1, t_2, \alpha\right)\right), \quad \alpha \in (0, 1),
\end{equation}
where $(q_1, t_1)$ and $(q_2, t_2)$ are the rotations and translations of views 1 and 2, and $(q_v, t_v)$ represents the interpolated virtual view. Here, Slerp and lerp are spherical and linear interpolation functions, respectively, with $\alpha$ controlling the interpolation degree.
For each virtual camera, we optimize the parameters of the 3D Gaussians using the following objective function:
\begin{equation}
    \mathcal{L}_{virtual} = \lambda_{dist} \mathcal{L}_{dist} + \lambda_{TV} \mathcal{L}_{TV},
\end{equation}
which combines depth distortion and total variance.
$\mathcal{L}_{virtual}$ shares the same hyperparameters $\lambda_{dist}$ and $\lambda_{TV}$ as the training view loss in Equation \ref{eq:train_loss}.
The computations of $\mathcal{L}_{dist}$ and $\mathcal{L}_{TV}$ do not require ground truth, making them applicable for virtual views.
Our virtual camera-based depth regularization aims to refine geometry from different views to avoid over-fitting, making the reconstructed endoscopic scene robust when viewed from significantly different perspectives compared to the training views.

\subsection{Sparse Control Nodes based Efficient 3D Gaussian Simulation}
\label{sec simulation}
Simulating reconstructed 3D scenes is an important research topic.
PhysGaussian\cite{xie2023physgaussian} first integrates physical simulation with the 3DGS technique, enabling realistic scene simulations. It treats each 3D Gaussian primitive as a material point, iteratively updating the shape of the 3D Gaussians using the deformation gradient from the MPM. However, this approach becomes slow when the number of 3D Gaussians is large. To capture the detailed textures of endoscopic scenes, 3DGS tends to generate a vast number of 3D Gaussians, making direct application of PhysGaussian impractical for efficient simulation.
To enhance simulation performance, we propose a sparse control nodes-based 3D Gaussian simulation method. 
This method employs a small group of sparse control nodes to deform the entire 3D Gaussian field, significantly reducing computational workload.

Initially, we sample the reconstructed 3D Gaussians using the farthest point sampling (FPS) algorithm~\cite{qi2017pointnet++} to obtain sparse control nodes. 
These nodes are then sent to the MPM algorithm for simulation.
After each simulation step, the shapes--specifically, the covariance matrices of the 3D Gaussians--are deformed according to the simulation results.
For each 3D Gaussian \( G_j \), we use the K-Nearest Neighbor (KNN) algorithm to identify its \( K \) (= 4) neighboring control nodes, denoted as \( \{p_k|k \in \mathcal{N}_j\} \). 
The weight between \( G_j \) and \( p_k \) is computed as \( w_{jk} = \text{softmax}(d_{jk}) \). The position of Gaussian \( G_j \) is updated as follows:
\begin{align}
    \mu_j^t &= \sum\limits_{k \in \mathcal{N}_j}w_{jk}\left(R_k\left(\mu_j-p_k\right)+p_k^t\right),
\end{align}
where \( \mu_j \) and \( p_k \) are the initial positions of \( G_j \) and control node \( p_k \), and \( p_k^t \) is the updated position of the control nodes. The rotation matrix \( R_k \) is derived from the polar decomposition of the deformation gradient of \( p_k \): \( F_k = R_kP_k \). 
The deformed covariance matrix of Gaussian \( G_j \) is computed as:
\begin{align}
    \Sigma_j^t &= F_j^t\Sigma_jF_j^{tT},
\end{align}
where \( F_j^t \) is computed by weighting the deformation gradients of surrounding control nodes: \( F_j^t = \sum\limits_{k \in \mathcal{N}_j}w_{jk}F_k \). Subsequently, the simulation results can be rendered by passing the deformed 3D Gaussians to the rasterizer.


\section{Experiments}\label{sec:exp}
We evaluate our reconstruction and simulation pipeline on the SCARED dataset~\cite{scared_dataset} and the StereoMIS dataset~\cite{stereomis}.
The evaluation is divided into two parts. 
The first part assesses our reconstruction method, with both qualitative and quantitative results showing that our approach effectively reconstructs high-quality endoscopic scenes from very sparse views. 
We use PSNR, SSIM, and LPIPS to evaluate the quality of synthetic views, and depth RMSE to assess the geometric accuracy of the reconstructed scenes.
The second part focuses on simulation performance, demonstrating how our sparse control nodes-based simulation method significantly enhances simulation speed compared to the baseline. 
The experimental results strongly validate the effectiveness and efficiency of our proposed method.

\subsection{Implementation Details}

In our implementation, the hyperparameters \( \lambda_{depth} \), \( \lambda_{dist} \), and \( \lambda_{TV} \) are adjusted based on the scene. All endoscopic scenes in our experiments are optimized for 3000 iterations. We use DepthAnything-V2\cite{depth_anything_v2} to estimate depth maps of endoscopic images, and camera positions are estimated using COLMAP\cite{schoenberger2016sfm}.
Frequency regularization is applied during the training of our method to mitigate overfitting resulting from sparse views, which constrains the degree of Spherical Harmonics (SH) to 0.
We evaluate our method using three train-test ratios: (1) 1:49, (2) 1:99, and (3) only the first and last frames of the endoscopy video clip are used for training. 
In each iteration, we generate a virtual camera from two sampled training views for regularization.
For simulation, we perform 80 substeps per step, with each substep having a timestep of 0.0005 s. 
The Young's Modulus is set to 10 kPa during the simulation. 
And the user-defined external forces are applied to the soft tissue to simulate the manipulation of the surgical tool.
All experiments were conducted on a single GTX RTX 3090 GPU.

\subsection{Qualitative and Quantitative Results of Reconstruction}
\begin{figure}[t]
\centering
\includegraphics[width=\textwidth]{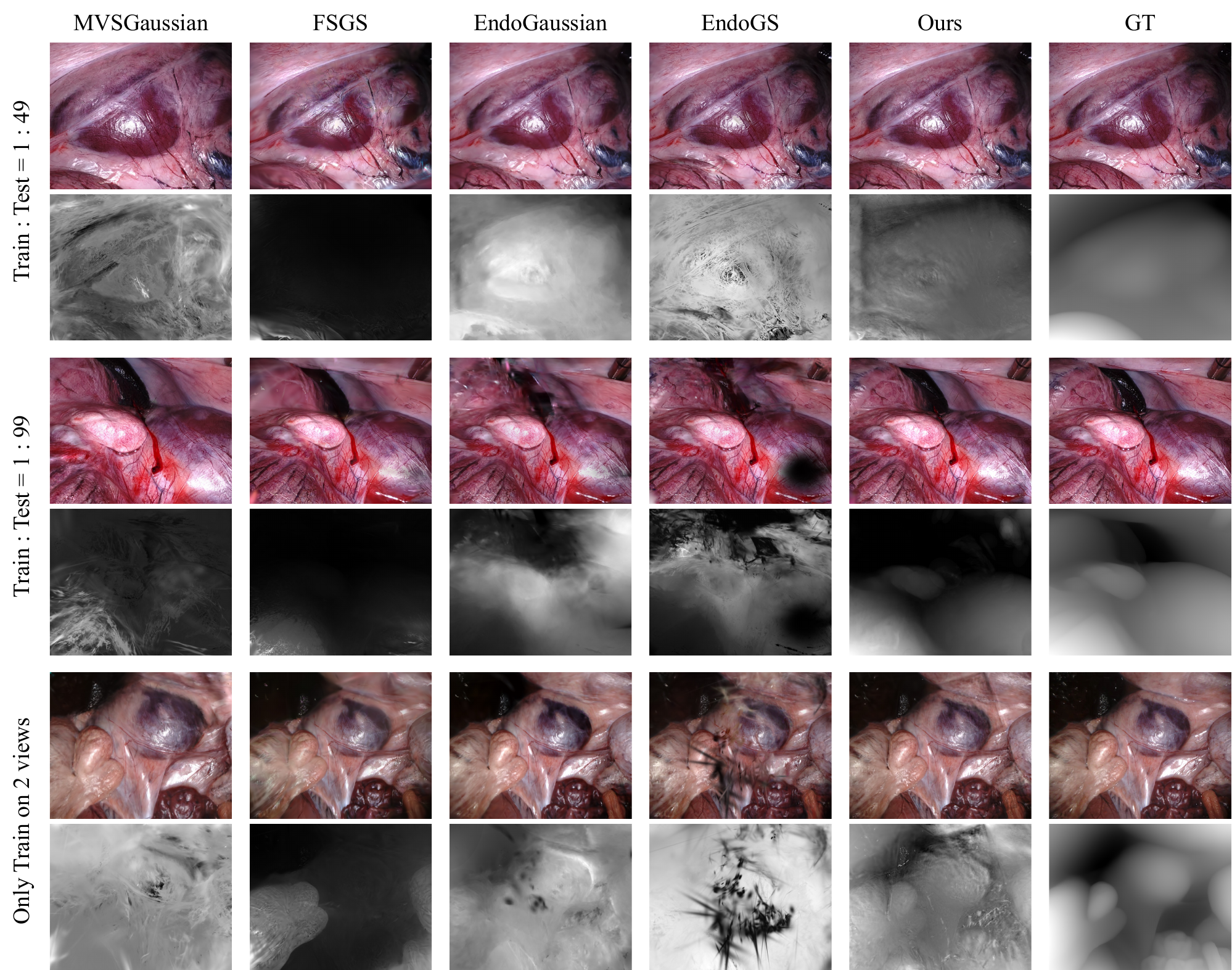}
\caption{Qualitative comparison with MVSGaussian\cite{mvsgs}, FSGS\cite{fsgs}, EndoGaussian\cite{liu2024endogaussian} and EndoGS\cite{zhu2024endogs} on rendered RGB images and depth maps on three train-test ratios.}
\label{fig:comparison}
\end{figure}

\begin{figure}[t]
\centering
\includegraphics[width=\textwidth]{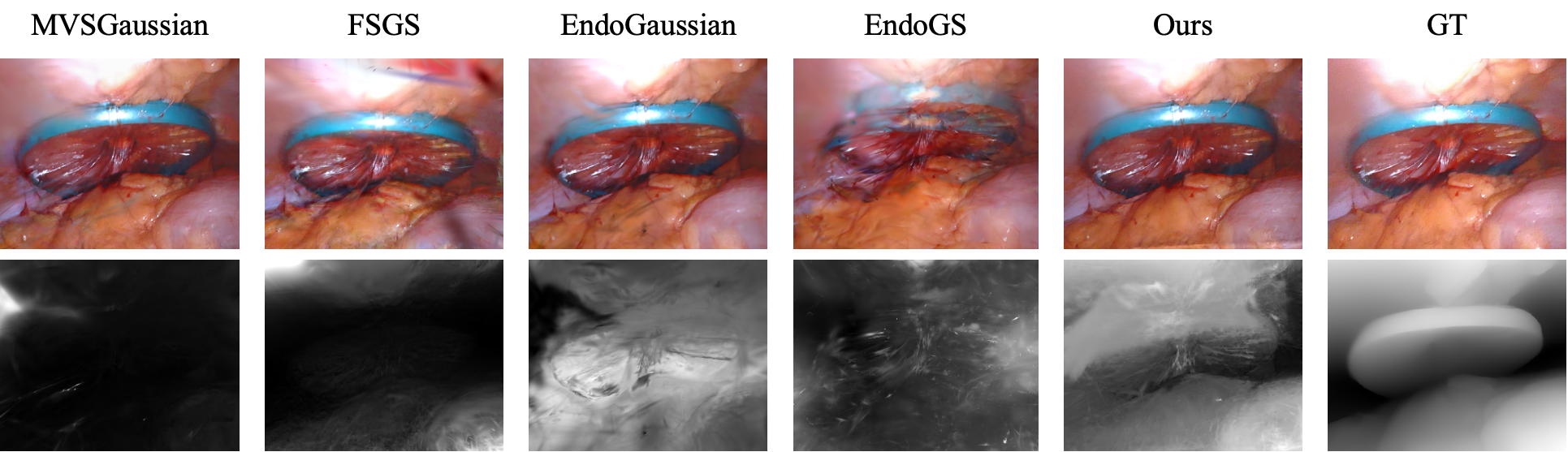}
\caption{
Qualitative comparison with MVSGaussian\cite{mvsgs}, FSGS\cite{fsgs}, EndoGaussian\cite{liu2024endogaussian} and EndoGS\cite{zhu2024endogs} on rendered RGB images and depth maps on StereoMIS~\cite{stereomis} Dataset.
}
\label{fig:comparison_mis}
\vspace{-1em}
\end{figure}

The evaluation is conducted across three train-test ratios. We compare our method with recent state-of-the-art sparse views reconstruction techniques\cite{fsgs, mvsgs} and competitive GS-based methods in the medical field\cite{zhu2024endogs, liu2024endogaussian}. The experimental results are listed in Table \ref{tab:comparison}.
Our method demonstrates superior performance across all train-test splits, validating the effectiveness of our proposed sparse view reconstruction approach. 
For a qualitative comparison, please refer to Fig. \ref{fig:comparison}. 
In the results, MVSGaussian\cite{mvsgs} fails to generate reasonable depth maps, while FSGS\cite{fsgs} produces overly smooth renderings and struggles with depth mapping at greater distances. EndoGaussian\cite{liu2024endogaussian} synthesizes unreasonable boundaries (notably in the lower left of the RGB result in the first row of Fig. \ref{fig:comparison}) and exhibits disorganized textures, as seen in the third row. EndoGS\cite{zhu2024endogs} tends to generate noticeable floaters under sparse training views, severely degrading synthetic view quality.
Further comparisons on the StereoMIS dataset are presented in Table~\ref{tab:comparison_mis} and Figure~\ref{fig:comparison_mis}.
In contrast, our reconstruction method effectively handles sparse view reconstruction, achieving detailed texture and reasonable depth maps simultaneously.
We also conducted an ablation study and present the experimental results in Table \ref{tab:ablation}. 
A performance drop is observed after removing any of these components,  which demonstrates that every part of our method contributes to the overall performance.

\subsection{Performance Improvement in Simulation}
\begin{figure}[t]
\centering
\includegraphics[width=\textwidth]{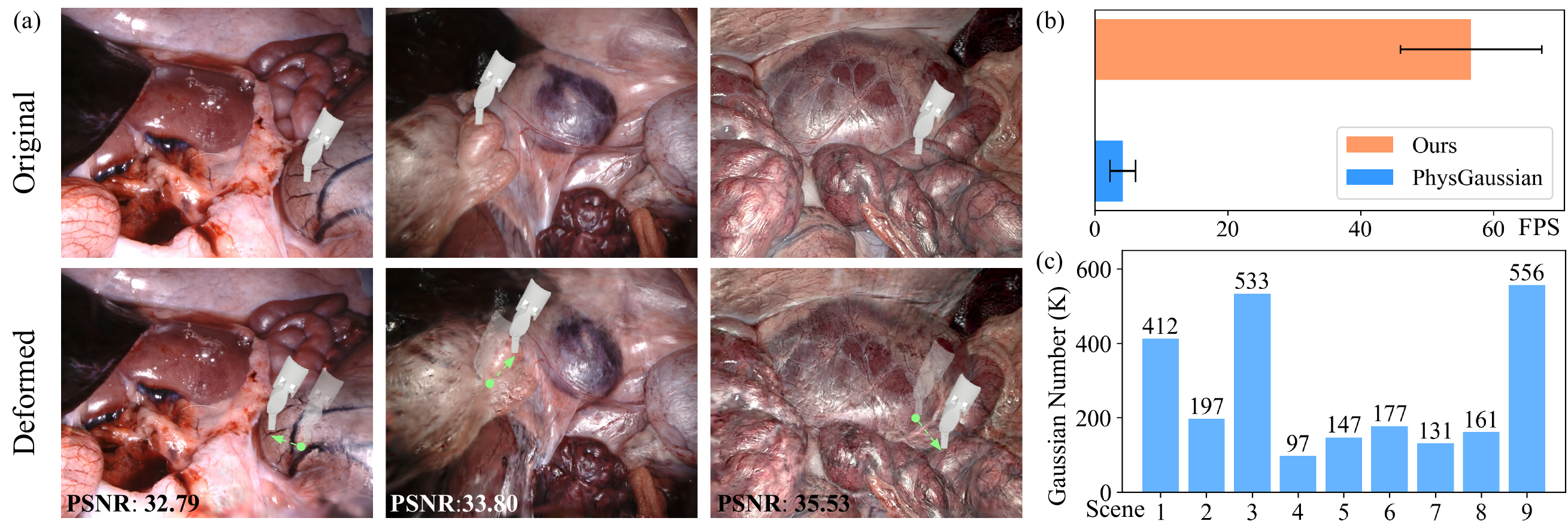}
\caption{\textbf{Qualitative and quantitative results of endoscopic scene simulation.} 
(a) Visualization of endoscopic scene deformation, with the position and direction of external forces indicated by green arrows. 
(b) Comparison of simulation frames per second (FPS) with PhysGaussian\cite{xie2023physgaussian}. 
(c) Number of Gaussians in each scene.}
\label{fig:deformation}
\vspace{-1em}
\end{figure}

As described in Section \ref{sec simulation}, we propose a sparse control node-based method to enhance the simulation speed of Gaussian Splatting. We conduct simulations on nine reconstructed endoscopic scenes using PhysGaussian\cite{xie2023physgaussian} and our method. 
The simulation implementation in SimEndoGS\cite{simendogs} is identical to that of PhysGaussian; therefore, we will not include a comparison with it.
As shown in Figure \ref{fig:deformation}(a), our method produces natural and physically plausible deformations under the manipulation of a surgical tool. The PSNR metric evaluates the differences between our method and PhysGaussian. PhysGaussian treats each Gaussian primitive as a material point, significantly increasing the computational burden when the number of Gaussians is large. In contrast, our method employs sparse control nodes to deform the entire endoscopic scene, greatly reducing computational workload and thus increasing simulation speed. 
As indicated in Figure \ref{fig:deformation}(b), our method achieves an average FPS of 57, which is approximately 14 times faster than the 4 FPS of PhysGaussian. All these experimental results demonstrate that our proposed method can effectively and efficiently perform endoscopic scene simulations.

\newpage
\begin{table*}[t!]
    \renewcommand{\tabcolsep}{3pt}
    \caption{
    Quantitative comparisons with five methods across three train-test ratios.
    }
    \centering
    \resizebox{\linewidth}{!}{
    \begin{tabular}{@{}l@{\,\,}|cccc|cccc|cccc|c}
    &   \multicolumn{4}{c|}{Train : Test = 1:49} & \multicolumn{4}{c|}{Train : Test = 1:99} & \multicolumn{4}{c|}{Only train on 2 views} & Train  \\
    & PSNR$\uparrow$ &SSIM$\uparrow$ &LPIPS $\downarrow$& RMSE $\downarrow$ & PSNR$\uparrow$ &SSIM$\uparrow$ &LPIPS $\downarrow$& RMSE $\downarrow$ & PSNR$\uparrow$ &SSIM$\uparrow$ &LPIPS $\downarrow$& RMSE $\downarrow$& Time \\ \hline
    EndoGaussian &25.44 &0.817 &0.382 &0.540 &23.52 &0.788 &0.398 &0.514 &20.32 &0.732 &0.417 &0.517 &3 min \\
EndoGS &\cellcolor{orange}26.42 &\cellcolor{yellow}0.840 &\cellcolor{yellow}0.283 &0.363 &23.87 &0.796 &\cellcolor{yellow}0.311 &0.377 &21.15 &0.742 &\cellcolor{yellow}0.350 &0.381&5 min \\
3DGS &\cellcolor{yellow}26.26 &\cellcolor{orange}0.847 &\cellcolor{orange}0.255 &\cellcolor{yellow}0.319 &23.44 &\cellcolor{yellow}0.801 &\cellcolor{orange}0.285 &\cellcolor{yellow}0.337 &20.82 &\cellcolor{orange}0.743 &\cellcolor{orange}0.325 &\cellcolor{yellow}0.317 &6 min\\
MVSGaussian &25.67 &0.758 &0.334 &0.320 &\cellcolor{yellow}24.06 &0.707 &0.363 &0.342 &\cellcolor{orange}22.20 &0.648 &0.404 &0.366 &2 min\\
FSGS &26.18 &0.836 &0.343 &\cellcolor{orange}0.261 &\cellcolor{orange}24.28 &\cellcolor{orange}0.807 &0.352 &\cellcolor{orange}0.273 &\cellcolor{yellow}22.07 &\cellcolor{tablered}0.769 &0.364 &\cellcolor{orange}0.269 &4 min\\
\hline
Ours &\cellcolor{tablered}27.19 &\cellcolor{tablered}0.855 &\cellcolor{tablered}0.254 &\cellcolor{tablered}0.143 &\cellcolor{tablered}25.28 &\cellcolor{tablered}0.819 &\cellcolor{tablered}0.279 &\cellcolor{tablered}0.137 &\cellcolor{tablered}22.94 &\cellcolor{tablered}0.769 &\cellcolor{tablered}0.307 &\cellcolor{tablered}0.184 &\hspace{0.3em}2 min
    \end{tabular}
    }
    \label{tab:comparison}
    \vspace{-1em}
\end{table*}
\newpage

\begin{table*}[t!]
    \renewcommand{\tabcolsep}{3pt}
    \caption{
    Quantitative results of ablation study.
    }
    \centering
    \resizebox{\linewidth}{!}{
    \begin{tabular}{@{}l@{\,\,}|cccc|cccc|cccc}
    &   \multicolumn{4}{c|}{Train : Test = 1:49} & \multicolumn{4}{c|}{Train : Test = 1:99} & \multicolumn{4}{c}{Only train on 2 views}  \\
    & PSNR$\uparrow$ &SSIM$\uparrow$ &LPIPS $\downarrow$& RMSE $\downarrow$ & PSNR$\uparrow$ &SSIM$\uparrow$ &LPIPS $\downarrow$& RMSE $\downarrow$ & PSNR$\uparrow$ &SSIM$\uparrow$ &LPIPS $\downarrow$& RMSE $\downarrow$ \\ \hline
    w/o Virtual Camera &26.22 &0.848 &0.254 &0.333 &23.63 &0.803 &0.283 &0.325 &21.09 &0.745 &0.324 &0.320
\\
w/o Frequency Reg. &25.61 &0.822 &0.298 &0.216 &23.87 &0.794 &0.310 &0.251 &22.08 &0.750 &0.331 & 0.229
\\
w/o Depth Distortion &26.20 &0.847 & 0.255 &0.328 &23.67 &0.803 &0.283 &0.334 &21.27 &0.747 &0.322 &0.321
\\ 
w/o Depth TV Loss &26.02 &0.847 &0.257 &0.340 &23.88 &0.803 &0.283 &0.350 &21.09 &0.744 &0.325 &0.315
\\ 
Ours &\textbf{27.19} &\textbf{0.855} &\textbf{0.254} &\textbf{0.143} &\textbf{25.28} &\textbf{0.819} &\textbf{0.279} &\textbf{0.137} &\textbf{22.94} &\textbf{0.769} &\textbf{0.307} &\hspace{0.3em}\textbf{0.184}
    \end{tabular}
    }
    \label{tab:ablation}
    \vspace{-1em}
\end{table*}

\subsection{Discussion on Reconstruction Efficiency}
Given a group of unposed endoscopic views, we initially estimate the camera positions for each view using COLMAP. Subsequently, posed endoscopic views serve as input to our method, and we initialize our model with the sparse point cloud generated by COLMAP. The time required for this step typically ranges from 5 to 15 minutes. However, this step can be bypassed for data where camera poses were recorded during the surgery. With these posed endoscopic views, our method can efficiently reconstruct endoscopic scenes in approximately 2 minutes, as demonstrated in Table~\ref{tab:comparison}.
It is evident that COLMAP can hinder the overall performance of the reconstruction pipeline. Recent advancements in general 3D vision, such as DUSt3R~\cite{wang2024dust3r} and VGGT~\cite{wang2025vggt}, are exploring the use of large-scale datasets to train feedforward neural networks capable of directly predicting camera poses and depth maps given the RGB inputs, which is faster than COLMAP. In future work, we intend to integrate these approaches to further enhance the performance of our method.

\begin{table*}[t!]
    \renewcommand{\tabcolsep}{3pt}
    \caption{
    Quantitative comparisons on StereoMIS~\cite{stereomis} dataset.}
    \centering
    \resizebox{\linewidth}{!}{
    \begin{tabular}{@{}l@{\,\,}|cccc|cccc|cccc}
    &   \multicolumn{4}{c|}{Train : Test = 1:49} & \multicolumn{4}{c|}{Train : Test = 1:99} & \multicolumn{4}{c}{Only train on 2 views} \\
    & PSNR$\uparrow$ &SSIM$\uparrow$ &LPIPS $\downarrow$& RMSE $\downarrow$ & PSNR$\uparrow$ &SSIM$\uparrow$ &LPIPS $\downarrow$& RMSE $\downarrow$ & PSNR$\uparrow$ &SSIM$\uparrow$ &LPIPS $\downarrow$& RMSE $\downarrow$ \\ \hline
    EndoGaussian\cite{liu2024endogaussian} 
&\cellcolor{yellow}21.65 
&\cellcolor{yellow}0.779 
&\cellcolor{yellow}0.351 
&0.398 
&20.71 
&0.727 
&0.355 
&0.392 
&\cellcolor{yellow}19.08 
&\cellcolor{yellow}0.710 
&0.382 
&0.399 
\\
EndoGS\cite{zhu2024endogs} 
&19.31
&0.676 
&0.430 
&0.463 
&17.18
&0.627 
&0.459 
&0.341 
&17.45
&0.628 
&0.441 
&0.364 
\\
3DGS\cite{kerbl20233d} 
&\cellcolor{orange}25.51
&\cellcolor{orange}0.800 
&\cellcolor{orange}0.279 
&\cellcolor{yellow}0.337 
&\cellcolor{orange}23.31
&\cellcolor{orange}0.738 
&\cellcolor{orange}0.305 
&\cellcolor{orange}0.305 
&\cellcolor{orange}22.06 
&\cellcolor{orange}0.727 
&\cellcolor{orange}0.315 
&\cellcolor{yellow}0.304 
\\
MVSGaussian\cite{mvsgs} 
&19.95
&0.734 
&0.370 
&0.452
&18.69
&0.690 
&0.403 
&0.363
&17.90
&0.669 
&0.414 
&0.352
\\
FSGS\cite{fsgs} 
&21.28
&0.736 
&0.402 
&\cellcolor{orange}0.325
&20.80
&0.709 
&0.364 
& \cellcolor{yellow}0.309
&18.91
&0.671 
&\cellcolor{yellow}0.377  
&\cellcolor{orange}0.303
\\
\hline
Ours &\cellcolor{tablered}27.23  &\cellcolor{tablered}0.816 &\cellcolor{tablered}0.261 &\cellcolor{tablered}0.170  &\cellcolor{tablered}24.20 &\cellcolor{tablered}0.743 &\cellcolor{tablered}0.300 &\cellcolor{tablered}0.222&\cellcolor{tablered}23.39 &\cellcolor{tablered}0.728 &\cellcolor{tablered}0.309 &\cellcolor{tablered}0.261

    \end{tabular}
    }
    \label{tab:comparison_mis}
    \vspace{-1em}
\end{table*}

\section{Conclusion}
This paper introduces an effective framework based on 3D Gaussian Splatting for surgical scene reconstruction from sparse endoscopy views and efficient, physically-based endoscopic scene simulation. 
We consistently use a 3D Gaussian representation for both reconstruction and simulation, allowing for convenient simulation, efficient visualization, and realistic visual results.
Our method features a specifically designed optimization strategy to reconstruct robust and accurate 3D endoscopy scenes from sparse images. The proposed simulation using sparse control nodes significantly reduces computational workload, enabling real-time simulation with user-defined interaction.
We hope that our method will inspire the development of interactive and highly realistic surgical scene generation, benefiting surgical training and surgical embodied intelligence.








\bibliography{wileyNJD-AMA}

\end{document}